# New Approaches for Natural Language Understanding based on the Idea -- that Natural Language encodes both Information and its Processing Procedures

Limin Zhang
E-mail: liminzhang08@qq.com

*Abstract* — we must recognize that natural language is a way of information encoding, and it encodes not only the information but also the procedures for how information is processed. To understand natural language, the same as we conceive and design computer languages, the first step is to separate information (or data) and the processing procedures of information (or data). In natural language, some processing procedures of data are encoded directly as the structure chunk and the pointer chunk (this paper has reclassified lexical chunks as the data chunk, structure chunk, and the pointer chunk); some processing procedures of data imply in sentences structures; some requests of processing procedures are expressed by information senders and processed by information receivers. For the data parts, the classification encoding system of attribute information and the information organization architecture (including constitutional structures of information sets and the hierarchy between the information sets) were discussed.

In section III, the theoretical part elaborated in section II has been verified in examples and proofed that the studies in this paper have achieved the goal of enabling machines to understand the information conveyed in the dialogue. In section IV, the author summarizes the basic conditions of "Understanding", rethinks what "Understanding" is and how to proceed.

The study in this paper provides a practical, theoretical basis and research methods for NLU. It also can be applied in large-scale and multi-type information processing in the artificial intelligence (AI) area.

## I. Relations between Information, Real world, and Natural Language

### A. Introduction

Natural language processing (NLP) techniques based on statistical models have achieved great success in machine translation. However, we are still far from letting machines understand natural languages, even for the simplest words: "Apple".

Unlike most previous approaches of NLP that focused on the structure study of words in sentences and context, this paper goes deeper to study the information represented by natural language. At the very beginning, the author tries to find out how human beings relate the text "Apple" to the physical "Apple". Inspired by the elementary information perception, transformation, and processing mechanisms in Neuroscience [1], the author discovers that human beings perceive the color, shape, smell, taste, and other information of physical "Apple" through their sensory systems; in brains, all this attribute information related to physical "Apple", such as color, shape, smell, taste, etc. forms an information set. Then, this information set can be encoded into texts, such as: "Apple", "苹果", "かんち", etc., so that, the understanding of text "Apple" is to understand the information set represented by text "Apple". Likewise, the information in the set represented by text "Apple" can also be encoded into texts (e.g. Color: red; Taste: sweet; Shape: round, etc.). As a result, the author discovered that there are architecture structures between words which represent information at different abstraction levels, and the understanding process of natural languages occurs on the information level rather than the lexical level (or morpheme), the author then takes these as inspiration to carry out the studies.

This paper also inspired by the study on the relational model of data by E.F. Codd's [2]; and draws on a wide range of elementary theories, ideas, and thinking methods in the following disciplines: discrete mathematics [3], structure of computer programs [4], computer operating systems [5], computer architecture [6], introduction to algorithms [17], etc.

### B. Some key concepts and relations between them

Human beings perceive the world through information received by neural systems; this information is a tiny fraction of all the information in the universe. The processes of information identification, classification, memorization, analysis, abstraction, association, etc. are the component activities of human thinking. To make the discourse more easily in section II, we must first introduce some key concepts: entities, attribute information, attribute space, information set, information encoding, memory-sheet, and expound the relations between them.

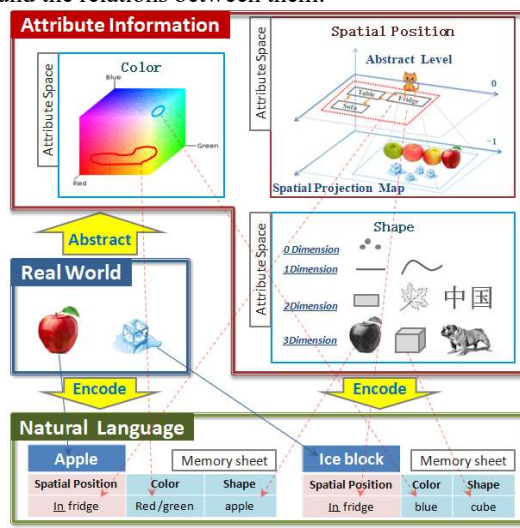

Figure 1. Three kinds of spaces are exhibit here, they are: Attribute Information Space, Real world, and Natural Language Space. The relations between them are highlighted in yellow arrows.

*Entities* in the real world can abstract out lots of different *attribute information*. Some are the basic attribute information sensed by human neural systems (e.g. vision, olfaction, gustation, audition, and somatic sensation, etc.), others are abstracted upon basic attribute information (e.g. classification, movement, relations, preference degree, etc.) Meanwhile, the entity also is defined and represented by all its attribute information.

The attribute information which is abstracted from entities in the real world can be classified by its natural properties. *Attribute spaces (AS)* are used to group attribute information that has similar properties. AS are *dynamic information sets*, the elements (or attribute information) in it are strongly related to the personal experience and the cognitive basis. In the research of NLU, the understanding of the data structure and the related algorithms of specific AS are the keys of NLU, which allow people to understand the changing-rules and changing-ranges of entities' attributes.

As illustrated in Fig.1, both entities in real world and attribute information in *attribute spaces (AS)* can be encoded into text information (natural language can be speech or text, in this paper, we only discuss the text.) *Memory-sheet* (in Fig.1) is a data structure that stores the entity word (text encoding of entities in real world) and its attribute words and phrases (text encoding of attribute information) together. All the attribute words and phrases in a Memory-sheet form a set, these attribute words and phrases are the elements of the set; the entity word can be seen as the representation or the name of the set. Thus, a Memory-sheet can display the mathematic relations between the entity and its attribute information, Entity A can be written as:

*Entity A = { $a_i$ | $a_i$ is an attribute of Entity A, i is a natural number}, $a_i \in A$.*

Up to now, we have seen attribute, attribute information, attribute words and phrases. The easy way to distinguish them is (i) when we talk about the attribute information of a specified entity, we use "attribute", in this case, we emphasized that the attribute is a part of the entity and should not be considered as independent information; (ii) when we talk about attribute information in AS, we use "attribute information" to emphasize its independence as the objectively existing information; (iii) for the attribute words and phrases, they are the text codings of attribute and attribute information, and won't reflect its independence.

Actually, people understand an entity by understanding its attributes. The more attribute are known, the better the entity is understood. The changes of an entity essentially are the changes of its attributes. Thus, when we say: "Give me an apple.", in essence, it is a request to change the spatial-position attribute of the apple. The verb "give" is a text encoding that represents change features of a sequence spatial-position attribute information.

Words in natural language are symbols that encode information. Besides the above entity words, attribute words or phrases, and changing feature words (verbs), we also can find the measuring words, interrogative, preposition, conjunction, punctuations, and specific sentence structure in natural language. Thus, understanding the information carried by words, punctuations, and specific sentence structures; distinguishing the function of the words, punctuations, and sentence structures in information transmission is the fundamental work in NLU.

*C. The relation between Natural Language and Information*

Let's think again about the relation between natural language and information. Natural language is a tool that human beings use to communicate with the outside world; it is also one of the carriers of information. Information constantly changes its carriers (or forms) in the process of transmission and processing, (i) in real world, information exists in the forms of electromagnetic wave, chemical molecules, and ions, the kinetic energy of air, etc., human beings perceive this information and transform them into biochemical and bioelectricity signals then process them in brains [1]; (ii) when people communicate with the outside world, information is then transformed into the form of natural language, body gesture, body movements, etc.; (iii) in CPU, information is processed in form of binary code. Obviously, the kinds and density of information contained in natural language are much higher than in other forms. Natural language has highly abstracted and conceptualized the information. Therefore, how natural language abstracts and conceptualizes information is the essential problem of NLU.

II. NEW CLASSIFICATION OF LEXICAL CHUNKS AND INFORMATION ORGANIZATION ARCHITECTURE

According to the grammar function, morphological standard, and the meaning standard, modern Chinese vocabulary has been divided into nouns, pronouns, verbs, adjectives, adverbs, prepositions, quantifiers, onomatopoeia, etc. [8]. On this basis, the author dismantles and analyzes the information carried by each type of lexical chunks, then reclassifying them into the data chunk, structure chunk, pointer chunk, and task chunk according to their functions and the roles they played in information transmission. See Fig.2, the task chunk is composed of data chunks, structure chunks, and pointer chunks.

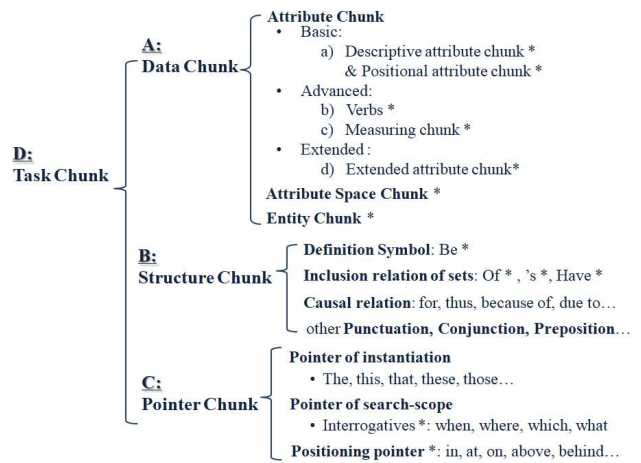

Figure 2.  New classification of lexical chunks

Due to the constraint of capability and time, the author cannot list all categories, items, and their usage scenarios, but only present the approach for reference and discussion. The items marked with an asterisk in Fig.2 are expounded in more detail below.

*A. Data Chunk*

As the name implies, data chunk plays the role of data in natural language, thus, the data structures of data chunks are the focus of NLU study. The classification of data chunks in Fig.2 is structurally based, for the sake of understanding, we first understand each sub-classification of data chunks through examples, and then elaborate them on the perspective of structure. There are hierarchical structures between data chunks; the concept of elements and sets in discrete mathematics can perfectly express this kind of hierarchical relationship, which also will be discussed.

*1) Attribute Chunk:* Attribute chunk can be words or phrases, they represent the attribute information which abstracted from entities. According to the different abstraction levels and methods, they can be further divided into the following four types.

*a) Descriptive Attribute Chunk and Positional Attribute Chunk:* Descriptive attribute chunks are the text encodings of the most basal kinds of attribute information perceived by *human neural systems*. The author prefers to classify the descriptive attribute chunks based on the study of the basic information perception system in the neurobiology of the brain, and they are the vision system (color, shape), the chemical senses system (taste, or gustation, and smell, or olfaction), the auditory and vestibular system, and the somatic sensory system [1]. It is easy to classify the related attribute words to the above sensory systems, see examples in braces below:

- <u>Color</u> {red, blue, green, orange},
- <u>Taste</u> {sweet, sour, acerb, bitter},
- <u>Shape</u> {square, round, cubic},
- <u>Smell</u> {smelly, balmy, pungent, apple-flavored},
- <u>Somatic sensation</u> {smooth, soft, furry}, etc.

The positional attribute chunks encode the spatial and temporal attributes of entities and usually appear in sentences in the form of phrases. (e.g. on this afternoon, in 1949, in the fridge, at school, etc.) The reason why this paper classifies positional attribute information as basic attribute information is that positional attribute chunks together with descriptive attribute chunks mapping the basic information dimension that humans needed to understand the world (time, space, and matter.)

*b) Verb:* Verbs are text encodings of change features which abstract from sequences of attribute information change records. Finally, the abstracted change features are recorded and encode as verbs in the memory, but the corresponding sequences of change records won't be recorded. Thus, this type of set is *called change features set*.

- *Fall* – represent the change features which abstract from sequences of spatial-position attribute information in spatial AS.
- *Sweeten* – represent the change features that abstract from sequences of taste attribute information in taste AS.
- *Run* – represent the change features that abstract from sequences of spatial-position information, distance attribute information, body posture information, etc. in corresponding AS.

*c) Measuring Chunk:* Measuring chunk is the text encoding of the reclassification result of the selected attribute information clusters. There are various methods and standards to implement the reclassify action: according to the different frames of reference, it can be divided into the *subjective measuring* and the *objective measuring,* according to the number of the measuring dimensions (or AS), it can be divided into the *single-dimension measuring* and the *multi-dimension measuring,* for example, speed is a *multi-dimension measuring* which needs to measure both the distance and the duration of time of a movement at the same time, and both the measuring of a distance and duration of time are the *single-dimension measuring*. And if the measuring result only has two values, such as "like" and "dislike", "agree" and "not agree", "yes" and "no", they can be called the *binary measuring*; if the measuring result has multiple values, such as "good", "better", "best" and "fast", "faster", "fastest", they can be called the *distribution measuring*. Some examples are given below for better understanding.

- *Distribution measuring:* Describe the distribution area of target objects after performing the statistical analysis on the attribute information in the selected measuring area. A distribution model is given in Table I to describe the data distribution, and each distribution area has the corresponding measuring words to describe it. Some examples of distribution measuring words are given in Table I.
- *Subjective measuring:* This type of measuring is adopted unified measuring standard to minimize the recognition tolerance of the same thing between individuals. E.g., Area (km2, m2), Speed (m/s, km/h), Temperature (°C, °F), Weight (g, kg, ton), Pressure (Pa), etc.
- *Quantity/Order/Ranking measuring:* We assume that the quantity measuring is based on the shape and spatial attribute information; the order measuring is based on the quantity, time, spatial-position, and other attribute information; and the ranking measuring is based on the order and other attribute information. It can be seen that the measuring dimensions of these three measurings are gradually superimposed and increased.

*d) Extended Attribute Chunk:* The extended attributes of an object are not defined by their content, but depend on their structural relations with the object. Extended attribute chunks are the text encodings of information chunks that have a specific structural association with the target object and will be elaborate in the information organization architecture section.

TABLE I. EXAMPLES OF THE DISTRIBUTION MEASURING CHUNKS

| Normal Distribution Model | Measuring Chunks | | Attribute Spaces (AS) being Measured | | |
|---|---|---|---|---|---|
| | | | Volume | Speed (distance + time) | Temperature |
| | never, beyond | | never seen, beyond the limit | | |
| | extremely, very, -est | | biggest | fastest | very hot |
| | Normal | a little bit, -er | bigger | faster | hot |
| | | average, proper | Average size | Proper speed | warm cool |
| | | a little bit, -er | smaller | slower | cold |
| | extremely, very, -est | | smallest | slowest | extremely cold |
| | never, beyond | | never heard, beyond the cognitive | | |

μ: Expected value
σ: Standard variation

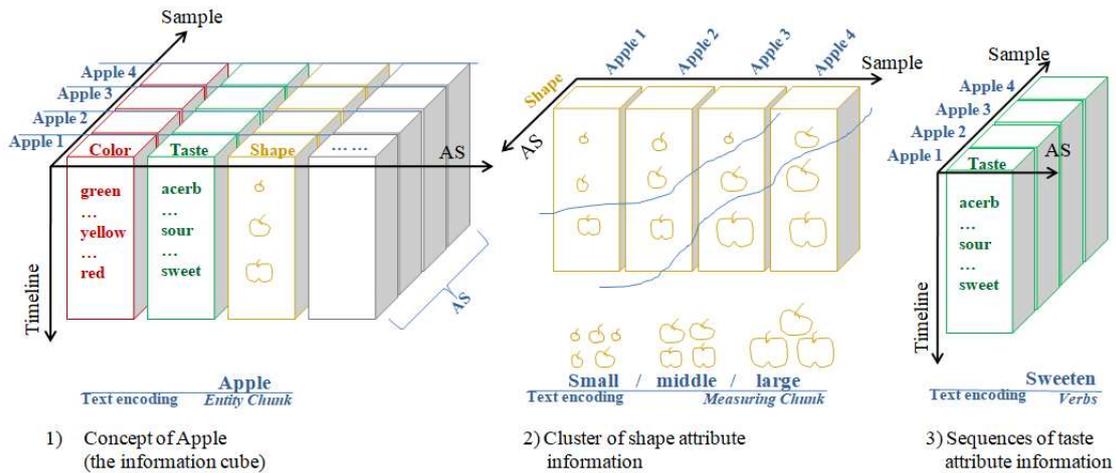

Figure 3. Structures of different attribute information sets

2) *Attribute Space Chunks (ASC)*: ASC are the text encoding of the AS, the underlined words outside braces of the examples in the attribute chunk section are ASC. Modeling the AS is crucial to NLU, because, for the understanding of attribute information, in addition to identifying which AS they belong to, we also need to know how this AS organizes and operates its attribute information, the variation-rules of these attribute information, and the predictions of the variation-boundaries of these attribute information in the AS, etc. Each AS is a independent subsystem of the whole thinking. In section III, the author models a spatial-positioning AS to help understand the concept of the AS more intuitively.

3) *Classification Encoding System of Attribute Information:* Looking back at all the above encoding laws of attribute information, we can discover that the attribute information is not independent and disorderly, but organized together in some forms, there is a classification encoding systems of the attribute information in our brains. The author would like to use figures (in Fig.3) to illustrate the constitutive modes of different attribute information sets. In the first figure, the descriptive attribute of samples are directly encoded as the basic attribute chunks and recorded in the corresponding AS. In the second figure, the clustering of the selected sample attribute information is been divided into three child clusterings according to their size, and the measuring chunks encode these three child clusterings as small, middle, and large, which are the indirect encoding of the selected sample attribute information clustering. In the third figure, the sequences of attribute information in taste AS are been selected, and its change feature are been encoded as sweeten (a verb), this also is an indirect encoding of the selected sample attribute information sequence. If we observe the usage of natural language, it is not hard to discover that the selected sample attribute information clusterings or sequences can be the basic attribute information, the extended attribute information, the advanced attribute information, or a mix of them. In the study of advanced attribute chunks, the coverage scope of the attribute information abstracted by the specific advanced attribute chunk is the key content of the study.

TABLE II. THE ENCODING COMPARISON OF COLOR ATTRIBUTE INFORMATION BETWEEN ENGLISH AND MODERN CHINESE

| | Original Classification | | New Classification | | |
|---|---|---|---|---|---|
| | *Noun* | *Adjective* | *Descriptive Attribute Chunk* | *ASC* | *Structure Chunk* |
| 🟥 | Red 红色 | Red 红色的 | Red 红 | color 色 | 的 [a] |
| 🟩 | Green 绿色 | Green 绿色的 | Green 绿 | color 色 | 的 |
| 🟧 | Orange 桔色 | Orange 桔色的 | Orange 桔 | color 色 | 的 |

a. Structrue chunks are highlighted in blue.

If compare several natural languages, we can discover that the encoding granularity and encoding overlap degree of the attribute information vary greatly in different natural languages. See examples given in Table II, for the color AS, the English text encoding is "color", and the Modern Chinese text encoding is "色", the English text encodings of the attribute information in color AS are "red", "green" and "orange", but these three words connotate the text encoding of color AS (color), which we think there is overlapping phenomenon. In Modern Chinese, text encoding of the attribute information in color AS are "红", "绿", "桔", during use, we can simply read the text encodings of the attribute information and the text encoding of the AS which they belong to, and put them together to describe the attribute information in the corresponding AS. When describing the inclusion relation between an entity and its attributes, simply add a structure word "的" between them, which will be elaborated in the structure chunk section.

We also can find out that, despite the above differences in encoding granularity and degree of overlap, but the overall encoding architectures are roughly the same. Perhaps we can ascribe this phenomenon to the common brain physiological structure human beings shared. And the application of this classification encoding systems in natural language greatly reduces the number of words and improves the expression efficiency.

4) *Entity Chunks (EC):* EC are the text encoding of the concept of entities in the real world. As we can see in Fig. 3, the EC "Apple" is the text encoding of the information cube. The operation of categorizing the attribute of the samples of physical apples and storing them in corresponding ASs in certain ways which form an information cube is the conceptualization of the physical apples. These conceptualization (or informatization) processes also are the processes of perceiving, classifying, and storing information from the environment by the nervous system. In natural language, there are three main methods to instantiate concepts, which will be elaborate in the pointer chunk section.

EC can be divided into the explicit EC and the implicit EC, according to whether they can be perceived by the visual system, some examples are given in Table III to help the understanding. The Explicit EC can be further divided into the dynamic EC and the static EC, according to the obvious difference in their spatial-position attribute. In spatial-position AS, static ECs are used as anchors to build the relative reference coordinate system, which will be elaborated in section III.

TABLE III. CLASSIFICATION OF EC

| | Classification | | Examples |
|---|---|---|---|
| Entity Chunk | Explicit EC | Dynamic EC | • *Human being, cat, car, cloud*<br>• *Apple, bag, laptop, cup* |
| | | Static EC | • *Sofa, house, school, shopping mall* |
| | Implicit EC | | • *Protein, carbohydrate, oxygen* |

Identifying the EC in a sentence is crucial to understanding the sentence. Because, all information conveyed in a sentence revolves around the EC, no matter the target of the sentence is to convey the information or to request an action. But the use of polysemous words in natural languages makes it difficult to accurately identify the information set encoded by the EC, for example, "Apple" is a kind of fruit, it also represents a smartphone. We can observe a large number of *lieu representations* used in natural languages, which use the partial attributes of the whole to represent the whole.  for instance, a little child imitates the barking of dogs (which is one attribute of physical dogs) to represent the physical dog instead of the speech of physical dog, or to use the brand of a smartphone (the brand is the extended attribute of a smartphone) to represent the smartphone, such as, my Blackberry, my Nokia, etc. The *lieu representation* usage is particularly prominent in English, that a large number of English words are both nouns and adjectives. Therefore, when we using the preset classification method to classify words, often leading to ambiguity in understanding, because the text encodings and the information they represent are not one-to-one mappings. But, this problem can be solved if we know the information organization architecture and can use the other data chunks in a sentence to accurately classify and locate the target data chunk.

5) *Information Organization Architecture:* Sets are the elementary organization forms of information. The author will discuss the information organization architecture in terms of the constitutional structures of information sets and the hierarchy between the information sets.

a) *Structural classification of information sets represented by different data chunks:* Let's understand the constitutional structures of information sets in terms of the conceptualizing process of physical objects. The nervous system perceives basic attribute information from different physical samples, classifying and storing them into corresponding AS. In the information cube in Fig.3, the column along the sample axis is the intersection of attributes of informationalized samples, the row along the AS axis is the union set of different informationalized attributes of a sample. Then add the third axis: the timeline. The information cube which categorizes and records the informationalized attributes of the sample clustering in sample, AS, and timeline dimensions forms the concept of that sample clustering. Thus, the information sets represented by EC are the *union sets*, the instantialted EC also represents a *union set*. As we can see in Fig.3, it is easy to understand that the information set represented by ASC, verbs, and measuring chunk are the *intersection set*. This feature enables us to make a preliminary judgment of the information set when it is been read. In neurobiology, may be we can identify the type of the stimulus signal by observe whether the activated neurons are in the same brain zone or in different zones.

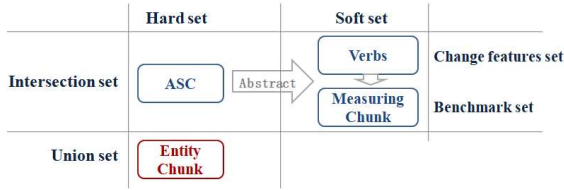

Figure 4. Structural classification of information sets represented by different data chunks

Looking at the encoding process of information sets, ASC and EC are the direct encoding of the information sets stored in memory database, but verbs and measuring chunks are the indirect encoding of the information sets, which need to implement extra algorithms on the selected information sets, then to encoding the results. Therefore, the author classify the information sets that represent by ASC and EC as the *hard set*, and the information sets represented by verbs and measuring chunk are the *soft set*.

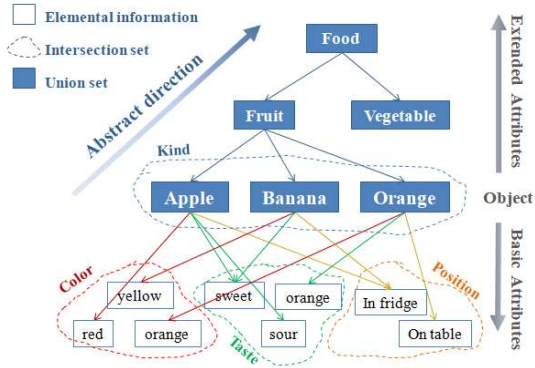

Figure 5. Memory-tree

b) *Hierarchical relation between data chunks*: The author believes there is no isolated information in brains, any information must be connected to other information, and the hierarchy is the endogenous structure of the natural language. The tree data structure can well express this hierarchical relation between data chunks. Observe the natural language, we can easily draw the memory-tree in Fig.

5. There are three kinds of vertices in the memory-tree, the elemental information which is represented by attribute chunks, the intersection sets which are represented by ASC, the union sets which are represented by EC. The arrows in the memory-tree are represented by structure chunks and will be expounded in the structure chunk section.

In the previous paragraph, we already discussed that the EC "Apple" represents the union set of all its attributes, such as red, sweet, sour, apple-shaped, etc. In Fig.5, the inclusion relationship between the EC "Apple" and its attributes is displayed as the vertex and its children vertices. Likewise, the vertices "Apple", "Banana", and "Orange" are the children vertices of the vertex "Fruit", thus, the EC "Fruit" represents the union of the "Apple" information cube, the "Banana" information cube, and the "Orange" information cube. By that analogy, we can continually do this abstraction (the union operation) again and again in higher layers of the memory-tree. And in a memory-tree, the higher the vertex, the bigger the information cube it represents and the larger density of information it contains. Or, we can say that the use of a higher abstract EC enables us to call information on a larger scale and requires better information processing capability. For instance: the EC "School" is the representation of the set {classrooms, playground, canteen, students, teachers, blackboards, chalks, books, courses, etc.} and the interactive activities between the elements in the above set. When we use the word "School," all its subsets and attribute information are been incorporated in.

The words "Apple", "Fruit" and "Food" in the memory-tree are all EC, but there is obvious hierarchical relationship between them, which can be written as: (Apple $\subset$ Fruit $\subset$ Food.)

TABLE IV.    EXAMPLES

| SN | Target Object | Attribute | | | | |
|----|---------------|-----------|---|---|---|---|
| | | Basic | | Advanced | | Extended |
| | | Descriptive | Positional | Measuring Chunk | Verb | Higher Layer EC |
| 1 | Apple | *is*[a] red | | | | . |
| 2 | Apple | *is* | in fridge | | | . |
| 3 | Apple | *is* | | bad | | . |
| 4 | Apple | *is* | | | roll | . |
| 5 | Apple | *is* | | | | **fruit** . |
| 6 | Apple | *is* | | | | **food** . |
| 7 | Fruit | *is* | | | | **food** . |

a. Structure chunks are highlighted in blue.

In natural language, we often describe an EC by using the data chunks which are associated with it. See sentences in Table IV, and ignore the missing articles and plural forms in these sentences, we take the EC "Apple" as the target object, and use the basic attribute chunks which is associate with and lower than "Apple" to describe the EC "Apple", then we get sentences 1 and 2. In sentences 3 and 4, the measuring chunk "bad" and the verb "roll" are not showing in the memory-tree, but they are associate with the EC "Apple", so we can use them to describe the EC "Apple". For the EC "Fruit" and "Food", in the memory-tree, they are associate with and higher than the EC "Apple", we also can use them to describe the EC "Apple", and get sentences 5 and 6. In a sentence, when the higher-positioned EC is used to describe the lower-positioned target EC, the higher-positioned EC is considered as an extended attribute of the lower-positioned EC. This explains why extended attributes are not defined by the content in particular, but by the structural relationship between them and the target description object.

The preset classification of data chunks allows for preliminary classification of the target data chunks, while further classification requires an auxiliary judgment from the information of the structural position of the target data chunk in the sentences. And the preset classifications of data chunks in this paper are also based on the compositional structure of the information sets they represent. Thus, the classification of data chunks is structural based discrimination.

The extended attributes of an EC increase as the connections between this EC and other higher-positioned EC increase. And the basic attributes of an EC are also expanded because of technological advances (e.g. the electron microscopy, the radio telescope, endoscopy, and MRI, etc.) In fact, the magnificent edifices of human thinking are built with these basic attributes information as bricks.

c) *Memory-graph:* Memory-graph is a cluster of memory-trees, and these memory-trees are connected in many ways. See the example of the memory-graph in Fig.6. When people build up their memory-graph, they take themselves as the center of this graph. In the process of growing and learning, people continually knit new information into their memory-graph by connecting the new information with the existing. Other important or close human beings can be set as the vice centers. These kinds of data structures which have one center and several vice centers are beneficial to improve the efficiency of search operations.

*Connections:* In a memory-graph, connections between data chunks are represented as the directed edges (arrows). There are two types of connections in Fig.6:

- Solid arrow: represent the real connection which implicates structural hierarchy, and always been used in memory-tree structures.
- Dashed arrow: represent the virtual connection that does not implicate structural hierarchy. The usage of a dashed arrow assumes that there is no hierarchical relation between different tree structures, so the dashed arrows are always been used to connect trees in a memory-graph (see subtree of "Cat" and subtree of "Dog" in Fig.6).

The above memory-tree and memory-graph are both data structures that can be used in a memory-database. Based on the above structures, the identification, classification, memorization, and even association operations of information are possible. However, all these operations are only related to the construction and the maintenance of the memory-database, which is also just a repository of information. And to understand natural language, we still need to build processing systems for information. Beside the AS which are used to process all kinds of attribute information, there still lots of high-leveled information processing systems, such as: decision system, motion control system, life support system, etc. The memory-database and information processing system can be considered as mutually independent systems. In section III, the author will introduce a spatial-position AS model and analyze its spatial information processing mechanism, to provide a reference for future study on information processing systems.

*B. Structure Chunk*

Connections between information sets can be interpreted as various kinds of relations, for example, the representation relation (defining relation), the inclusion relation, the causal relation, and so on. In this section, we will discuss the defining relation and inclusion relation represent by structure chunks: "Be," "Of," "'s" and "Have," and elaborate on two corresponding data reading modes: the defining reading mode and the set reading mode.

1) *"Be"*: In dictionaries, "Be" and "Have" are classified as verbs, which is against the verb classification rule introduced in the data chunk section. In natural language usage habits, we can observe that the data chunks after "Be" always are used to explain or define the data chunk before "Be". Although the author say that an entity is defined and represented by all its attribute information, but people do not need and impossible to completely describe an entity in course of natural language usage. Usually, people just partially describe an EC by giving one or several of its attributes. The given attributes are also used to help locate the EC. In natural language, the data chunks after "Be" are used to describe and define the data chunks before "Be".

2) *"Of", "'s" and "Have"*: These words interpret the connections as inclusion relations between data chunks. We assume that there are no equal sets in a memory-graph. So, the inclusion relation of sets can be written as:

- A ⊃ B → A **has** B. or A**'s** B.
- B ⊂ A → B **of** A.

Now, we can simulate the process of how brains read the data in memory-graph (in Fig.6) in below two modes, the read-out sentences are list in Table V.

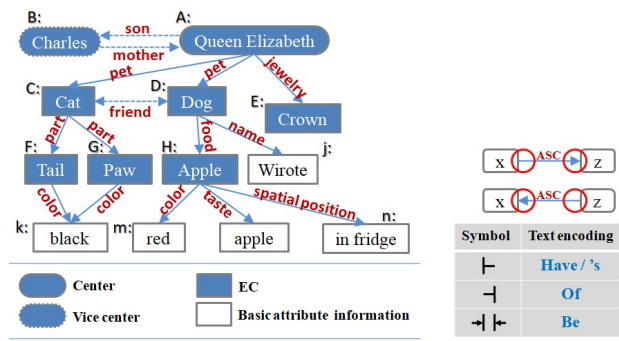

Figure 6.  Queen's memory-graph

a) *Defining Reading Mode (DRM):* Or we can call it the full reading mode, which read whole data chunks from a selected reading chunk. As we can see in Table V, though the article and punctuation are missing in those sentences, we still can roughly get the information they conveyed. In defining reading mode, if the following attribute chunk's connotation (the data chunk after "be") can cover the current ASC's, the ASC can be omitted in expression. It is rare to see this phenomenon in Modern Chinese because the classification coding system of Modern Chinese has smaller coding granularity and lower coding overlap compare to English.

TABLE V.   EXAMPLES OF THE DEFINING READING MODE AND THE SET READING MODE

| Data Reading Mode | | **Defining Reading Mode** (Full Reading Mode) | | | | **Set Reading Mode** (Inclusion relation Reading Mode) | | |
|---|---|---|---|---|---|---|---|---|
| Structural CLS [c] | | **Target Object Chunk** | ⊣⊢ | **Attribute Chunk** | | **Set** | ⊢ | **Subset** |
| Lexical CLS | | *Entities* | *ASC* | *Attribute Information* | *Entities* | *Entities* | *ASC* | *Entities* |
| Items | Reading Chunk | | | | | | | |
| 1 | H ASC m | Apple | *'s* | ~~color~~ [a] | *is* [b] red | Apple | *has* | color |
| 2 | H ASC n | Apple | *'s* | ~~spatial-position~~ | *is* in fridge | Apple | *has* | spatial-position |
| 3 | D ASC j | Dog | *'s* | name | *is* Wirote | Dog | *has* | name |
| 4 | [ F , G ] ASC k | Tail *and* paw | *'s* | ~~color~~ | *are* black | Tail *and* paw | *have* | color |
| 5 | A ASC E | Queen Elizabeth | *'s* | jewelry | *is* crown | Queen Elizabeth | *has* | jewelry |
| | | | | | | Queen Elizabeth | *has* | crown |
| 6 | A ASC [ C , D ] | Queen Elizabeth | *'s* | pet | *are* cat *and* dog | Queen Elizabeth | *has* | pet |
| | | | | | | Queen Elizabeth | *has* | cat *and* dog |
| 7a | A ASC1 B | Queen Elizabeth | *'s* | son | *is* Charles | Queen Elizabeth | *has* | son |
| 7b | A ASC2 B | Queen Elizabeth | | | *is* mother *of* Charles | | | |
| 8a | B ASC1 A | Charles | *'s* | mother | *is* Queen Elizabeth | Charles | *has* | mother |
| 8b | B ASC2 A | Charles | | | *is* son *of* Queen | | | |
| 9a | D ASC C | Dog | *'s* | friend | *is* cat | Dog | *has* | friend |
| 9b | D ASC C | Dog | | | *is* friend *of* cat | | | |
| 9c | C ASC D | Cat | *'s* | friend | *is* dog | Cat | *has* | friend |

a. If the following attribute chunk's connotation can cover the current ASC's, the ASC can be omitted in expression.
b. Structure chunks are highlighted in blue.
c. CLS is the abbreviation of classification.

b) *Set Reading Mode (SRM):* this is an inclusion relation reading mode, which only reads information sets (EC and ASC) and the inclusion relation between them from a selected reading chunk. We can find three typical usages in natural language:

- A ASC b : A ⊃ ASC, b∈ASC.
"b" is the elements of "ASC", and b is not a set, thus, only read the "A ⊃ ASC" part from " A ASC b ", and omit the "b∈ASC" part. E.g., items 1, 2, 3, and 4 in Table V.

- A ASC B : A ⊃ ASC ⊃ B.
In this case, we select A as the target description object, then, we can choose the "A ⊃ ASC" part, or the "A ⊃ B" part to read-out which is according to the requirement. E.g., items 5 and 6 in Table V.

- A ASC B : ASC ⊂ A, B ⊄ A.
Due to there is a virtual connection between A and B, structurally, there is no inclusion relation between A and B, just the virtual abstract relation. This virtual abstract relation can be unidirectional or bidirectional. In this case, we can choose the "ASC ⊂ A" part to read-out. E.g., items 7a, 8a, 9a, and 9c in Table V.

3) *Punctuations and conjunctions:* Punctuation and conjunctions segment the information on a larger scale, for the following purposes:

- Distinguish the task type of segmented information chunks (e.g., period, question marks, exclamation marks.)
- Distinguish the processing order of segmented information chunks (e.g., punctuations: commas, semicolons, parentheses.)
- Indicate structural relations of segmented information chunks (e.g., conjunctions: and, or, therefore, so.)

Except for structure chunks listed in Fig.2, sentence structures and paragraph structures segment information chunks on an even larger scale. The structural relations represented by those structural forms are also more diversified. Sentence structures will elaborate in the task chunk section.

*C. Pointer Chunk:*

  1) *Pointer of instantiation:* During use, the instantiation of a concept can be achieved by adding a pointer chunk "the" in front of the EC and giving characteristics that distinguish the entity from other entities of the same class. The instantiated EC are underlined and pointer chunks are hightlight in orange in below examples:
- **The** apple in the fridge.
- **The** apple with a scar.

Or directly point out the target object from existing entities by using pointer chunks (e.g. this, that, these, those, etc.)
- **This** cat.
- **That** house.
- **Those** people.

Or to provide the other EC which are already known and adjacent to it to help to locate it accurately in a memory-database, thereby instantiating the target EC.
- Her (she**'s**) dog.
- Queen Elizabeth**'s** dog.

  2) *Pointer of search-scope:* The interrogatives together with the AS in sentences are used to limit the area or scopes that should carry out the information search operation, therefore, they been called the pointer of search-scope in this paper.

|  | **Search scope** |
|---|---|
| **What color** is the apple? | AS of color |
| **Where** are we going? | AS of spatial-position |
| **When** shall we leave? | AS of timeline |
| **How old** are you? | AS of age |
| **How fast** the car is? | AS of speed measuring |

  3) *Positioning pointer:* Positioning pointers used upon different AS represent different kinds of positional information. In spatial-position AS, pointer chunks are used for positioning the target object in space. On the timeline, positioning pointers are used to positioning target objects on the timeline. Thus, the understanding of the positioning pointers is relying on the understanding of the characteristics of the corresponding AS, some examples are given below. And no matter the positioning of the target objects are in space or the temporal systems, the accuracy of positioning, to choose the relative-position positioning or the absolute-position positioning should be extraordinarily considered, which will not be expanded here.
- **in** the fridge / **on** the table / **at** home / **behind** the door
- **in** this morning / **on** Monday / **at** 6 clock / **after** that day

*D. Task Chunk:*

Task chunks usually are sentences which composed of data chunks, structure chunks, pointer chunks, and other lexical chunks to express specific task requirements (operations on data chunks). According to different sentence structures that express the specific type of task, sentences can be divided into the data description task, the data verification task, and the data search task. See Table VI, the above task types can be subdivided again according to different data reading modes.

TABLE VI. NEW CLASSIFICATIONS OF SENTENCES

| Data Reading Mode<br>Task Type | DRM | SRM | PRM |
|---|---|---|---|
| A. Data description task | A-DRM | A-SRM | A-PRM |
| B. Data verification task | B-DRM | B-SRM | B-PRM |
| C. Data search task | C-DRM | C-SRM | C-PRM |

Before introducing the classification of task chunks, the author would like to introduce another type of data reading mode: the attribute-changing process reading mode (PRM). As the name implies, PRM reads EC and their attribute-changing processes. Items 6, 7, and 8 in Table VII are the sentences read out in PRM, attribute-changing processes of entities in sentences are represented by verbs which are the change features abstract out from sequences of attribute information changing records. Attribute-changing processes can be interactive or noninteractive. Item 6 in Table VII is the description of the noninteractive attribute-changing process of the entity. Items 7 and 8 are the description of interactive attribute-changing processes between entities. In the interactive attribute-changing processes, it is important to distinguish the active role and the passive role. Usually, entities before verbs are active roles; entities after verbs are passive roles (the passive tense is not discussed in this paper).

Now, the author will put the information processing entity (IPE) in the natural language receiver's shoes, and then elaborate on the understanding process of each type of task chunks in Table VI. The scenarios that put IPE in the natural language sender's shoes won't be discussed in this paper.

  1) *Type A tasks:* When IPE gets information input of data description type tasks, they understand the input information by activating or mobilizing the data chunks described by input information in memory-database. The activating or mobilizing processes of the data chunks described by input information are also been considered as data reading operations in IPE's memory-database (the data reading operation is highlighted in bold in Fig.9.) The understanding of A-DRM and A-SRM types of sentences focus on the understanding of structural relations between data chunks in sentences. The understanding of the A-PRM type of sentences focuses on the understanding of attribute-changing progress of EC in sentences.

It is easy to find out that Imperative Sentences are the abridged expression of A-PRM type of sentences that request the IPE to take immediate action. Exclamatory Sentences are the abridged expression of data description type of sentences that emphasize the extraordinary attributes of target objects, the target objects are assumed to be known information by default and are omitted in the expression.

TABLE VII. EXAMPLES OF NEW SENTENCE CLASSIFICATIONS

| | | Column A | Column B | Column C |
|---|---|---|---|---|
| | Original CLS | **Declarative & Exclamatory Sentences** | **Interrogative Sentences** (Yes/No & choice question) | **Interrogative Sentences** (*WH-word question*) |
| | Task Type | **A: Data Description Task** | **B: Data Verification Task** | **C: Data Search Task** |
| Data Reading Mode | | Text- pointer words; Text- structure words; Text- measurment words; Text- verbs; | "Be", "Have" and verbs that need to be varified are marked in ▭. | "◤" Red triangles in column A mark out the missing data chunks and the reading direction. Text- Pointer chunk of search-scope |
| 1 | DRM | This apple *is* red. | Is this apple *is* red ? | What color ~~red~~ is this apple ? |
| 2a | DRM | The dog's name *is* Wirote. | Is the dog's name *is* Wirote ? | What ~~Wirote~~ is the name of the dog ? |
| 2b | | The dog's name *is* Wirote. | | Whose the dog's name is Wirote ? |
| 3a | DRM | They *are* Queen's crowns. | Are they *are* Queen's crowns ? | Whose ~~Queen's~~ crowns are they ? |
| 3b | | They *are* Queen's crowns. | | Which ~~they~~ are Queen's crowns ? |
| 4 | SRM | The cat *has* a black tail. | Does the cat ha~~s~~ve a black tail ? | Who ~~the cat~~ has a black tail ? |
| 5a | SRM | Queen *has* twelve crowns. | Does Queen ha~~s~~ve twelve crowns ? | Who ~~Queen~~ has twelve crowns ? |
| 5b | SRM | Queen *has* twelve crowns. | | How many ~~twelve~~ crowns **does** Queen ha~~s~~ve ? |
| 6 | PRM | Wirote run away. | Does Wirote run away ? | Who ~~Wirote~~ run away ? |
| 7 | PRM | Queen read the book. | Did Queen read the book ? | Which ~~the~~ book **did** Queen read ? |
| 8 | PRM | Queen likes coffee and tea. | Does Queen likes coffee and tea ? | What kind of drink ~~coffee and tea~~ **does** Queen likes ? |

Notes: Auxiliary words which highlight in yellow are still under study.

2) *Type B tasks:* Human beings do information recognition all the time while they are awake. Information recognition is the operation to compare input information with the existing information in their memory-database. When the input information is new and does not have a related record in IPE's memory-database, the IPE can complete this round of information recognition by calling someone else's memory-database.

In natural language, the information senders use the data verification task type of sentences to express data verification requests to the information receivers that request to invoke corresponding information in information receivers' memory-database to help to complete the data verification task. See Table VII, the sentences in column B express the verification request for information described in column A. Specific implementation methods are as follows:

a) *For the B-DRM type of sentences:* This type of sentences request to verify the "Be" structural relation between the input data chunks. The request is achieved by pop out the "Be" in verification data chunks and relocate it to the beginning of the data verification task chunks (sentences), and add question marks to end these data verification task chunks.

b) *For the B-SRM type of sentences:* This type of sentences request to verify the inclusion relation between the input data chunks. The request is achieved by add "Do" at the beginning of the data verification task chunks, and also add question marks to end these data verification task chunks.

We can find some clues in Fig.6 for why the extra auxiliary words "do" are needed in B-SRM type of task chunks. It is obvious that "Be" is a bidirectional structure relation, and "Have" is a unidirectional structure relation. Thus, "Have" needs to be kept between the set chunk and its subset chunk to indicate this unidirectional structure relation. Therefore, extra auxiliary words "Do" are added at the beginning of the data verification task chunks to express the verification requests of the inclusion relation.

c) *For the B-PRM type of sentences:* This type of sentences requests to verify the attribute-changing processes represented by verbs. Same as B-SRM type of sentences, the B-PRM type of task chunks express the attribute-changing process verification request by add "Do" at the beginning of data verification task chunks, and add question marks to end these data verification task chunks.

All attribute-changing processes are directional, so verbs also need to be kept in sentences and extra auxiliary words "Do" need to be added at the beginning of the data verification task chunks to express the verification requests of the attribute-changing process.

This paper won't discuss the usage of variants of the words "Be", "Have", "Do" and verbs. The classification of the auxiliary word "Do" is still under study.

3) *Type C tasks:* In the process of information processing, if some information is found missing, IPE can send requests to other IPE to assist in searching the missing information through the data search task chunks. In the data search task chunks, the missing information is substitute by the pointer chunks of search-scope (the red text in Table VII) which indicates the area or scopes where shall carry out the search operation.

In the data search task chunks, the pointer chunks of search-scope are always been placed at the beginning of the whole data search task chunks. Thus, when the missing part is at the lower-leveled position in a structural relation or the missing parts are the passive roles in attribute-changing processes, the reading order of the data chunk needs to be adjusted accordingly. The different scenarios are discussed below:

a) *For the C-DRM type of sentences:* Due to the "Be" structure relation is bidirectional, no matter the missing data chunk is before or after the "Be," take the missing data chunk as the start reading point, then to read the whole data chunks one by one. The readout task chunks list in column C and mark out with underlines. Of course, the missing data chunk needs to be substituted by the pointer chunk of search-scope, and a question mark is added to end this data search task chunk.

b) *For the C-SRM type of sentences:* "Have" represents the unidirectional structure relation, the data chunk behind "Have" is the subset of the data chunk which before "Have". Thus, when the missing data chunk is before "Have", the order of data chunks in the task chunks won't be changed (e.g. items 4 and 5a in Table VII); but when the structurally lower-leveled data chunk (the data chunk behind "Have") is the missing part, the missing data chunk still be set as the start reading point, but the "Have" need to be attached behind the higher-leveled data chunk (the data chunk before "have") to indicate its position in the unidirectional structure relation, extra auxiliary words "Do" is added between the lower-leveled data chunk and the higher-leveled data chunk to separate them (e.g. item 5b in Table VII), the readout sentences are listed in column C and mark out with underlines. Of course, the missing data chunk is substituted by the pointer chunk of search-scope, and a question mark is added to end this data search task chunk.

c) *For the C-PRM type of sentences:* Same as C-SRM type of task chunks, when the missing data chunk is the active role in the attribute-changing process, the order of data chunks in the task chunk won't be changed; but when the missing parts are the passive roles, the whole data search chunk needs to start with the missing passive part, and the verbs need to be attached behind the active role to indicate its active position, the extra auxiliary word "Do" is added between the active role and the passive role to separate them (e.g., item 7 and 8 in Table VII). Finally, a question mark is added to end this data search task chunk.

All the above studies do not involve tenses. The tense system in natural language is a description of the relative position relation between the current time position on the timeline and the time position when the data is recorded on the timeline.

So far, the author separates the information (data) and the processing procedure of information (data) in natural language and briefly introduced the research thinking of the information organization architecture and new classifications of lexical and sentence type. In the next section, the author takes the spatial-position attribute information as an example to illustrate how the spatial-position information is been organized and processed in spatial AS.

### III. SOME EXAMPLES

#### A. One Spatial-position Attribute Space Model

An attribute space (AS) that represents the spatial relations between entities in the real world is called spatial-position AS. In spatial-position AS, the spatial information is recognized by the human visual sensory system, then abstracted out the scope and direction relations between entities, and then stored these entities and their spatial scope and direction relations in the corresponding models. Before introducing the spatial projection map (SPM) which is one of the spatial-position AS models, two concepts need to be introduced:

TABLE VIII. SPATIAL-POSITIONING WORDS AND PHRASES

| Category | Examples of SPW | Symbol & Coordinate System |
|---|---|---|
| **Scope Relation Recognition** | | |
| *Inside* | In, at, inside, within, among | $\in$ , $\subset$ , $\supset$ |
| *Outside* | Out of, outside, beyond | $\notin$, $\not\subset$ |
| **Direction Relation Recognition** | | |
| • *Relative direction* | | |
| *Upper side* | on, above, up, over | |
| *Down side* | under, below, beneath | |
| *Front:* | before | |
| *Back:* | after, behind | |
| *Left side* | on the left side | |
| *Right side* | on the right side | |
| *Others* | against, toward | |
| • *Absolute direction* [a] | | |
| | East, west, south, north, middle | |
| **Distance Relation Recognition** | | |
| | By, beside, alongside, nearby, around, close to, next to | |

a. The absolute concept only exists within a specific range and scale, which won't be discussed in this paper

1) Spatial-positioning words (SPW) and space-assisted positioning points (SAPP):

a) *Spatial-positioning Words (SPW):* SPWs are used to represent the position information in spatial-position AS. There are not many words used for spatial-positioning in natural language, and according to different recognition and judging mechanisms, they can be divided into *the spatial scope relation recognition*, *the spatial direction relation recognition*, and *the spatial distance relation recognition*. According to the coordinate system used, the spatial direction recognition can be divided into *the relative direction recognition* and *the absolute direction recognition*, see the examples in Table VIII.

b) *Space-assisted Positioning Point (SAPP):* The static EW (see Table III) which used for space-assisted positioning purpose in natural language, are called SAPP in this paper.

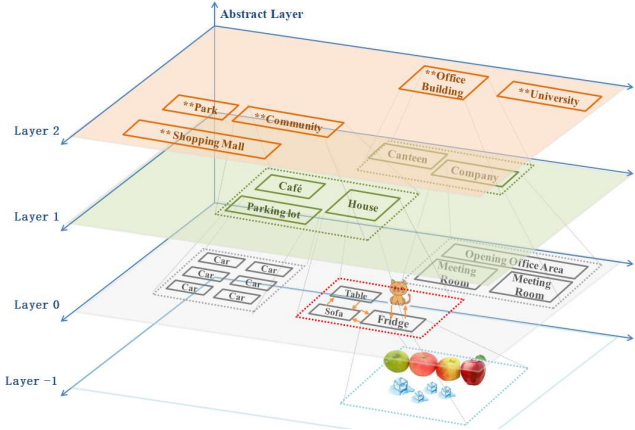

Figure 7. Graph-tree structure of SPM

2) *Spatial Projection Map (SPM):* Now, we will model the spatial relation between entities using SPW and SAPP. Spatial projection map (SPM) is a graph-tree model representing the scope and relative direction relations between relative static entities. Static entities are being abstracted out as vertices elements of the model. Scope and relative direction relations are being abstracted out as edges elements of the model. As we can see in Fig.7, SPM is a hybrid model combined with horizontal graph structures and vertical tree structures. The relative direction relations are stored in the horizontal graph structure, and scope relations are stored in the vertical tree structure. The author use the adjacency matrix to represent the graph structure of the SPM in Fig.7, details are as follows:

a) *Graph Structure Model (in-layer or horizontal structure):* The vertices or SAPPs are stored in graph structure according to their relative direction relations in the real world. They also can be represented in mathematic as below:

- Graph G=<V, E> consists of V, a set of all the SAPPs in the same layer called vertices, and E, a set of six fixed relative direction relation of V called edges.
- $G_i = <\bigcup_{j=1}^{n} G_i^j>$, i: serial number of layer, j: serial number of the subgraph.

TABLE IX. ADJACENCY MATRIX REPRESENTION OF SUBGRAPH $G_0^1 = <V_0^1, E>$

| SN | $V_0^1$ \ E | Left side | Right side | Front | Back | Upper side | Down side |
|---|---|---|---|---|---|---|---|
| 1 | Table | Φ | Φ | Φ | Sofa | Apple[a] | Φ |
| 2 | Fridge | Sofa | Φ | Φ | Φ | Cat[a] | Φ |
| 3 | Sofa | Φ | Fridge | Table | Φ | Φ | Φ |

a. "Apple" and "Cat" are Dynamic EW, thus, they are not been listed in $V_0^1$ set.

Take the red dotted box portion of Layer 0 in Fig.7 as an example, Table IX is a adjacency matrix which represents the subgraph $G_0^1 = <V_0^1, E>$. In the subgraph $G_0^1$, the vertices set $V_0^1$ = {table, fridge, sofa} and the edges set E = (left side, right side, front, back, upper side, downside), the elements in the vertices set can be added or subtracted according to the actual situation, but may not be repeated, the edges set is a tuple consisting of six fixed spatial direction relation. If vertices are not on these six directions, the spatial direction relation between the subject and object needs to be calculated, the calculation method won't be discussed here.

b) *Tree Structure Model (between-layer or vertical structure):* See Fig.8, the tree structure is the vertical structure of the graph-tree model, which is a between-layer structure. It represents the spatial scope relation of vertices between adjacent layers. A graph-tree model can consist of many trees. The root, leaves, and internal vertices of each tree can be distributed in different layers. The spatial scope of each vertex is represented by the union of the spatial scope of all its children vertices. See Fig.8, the spatial scope of the internal vertex "House" is the union of {Table, fridge, sofa …}.

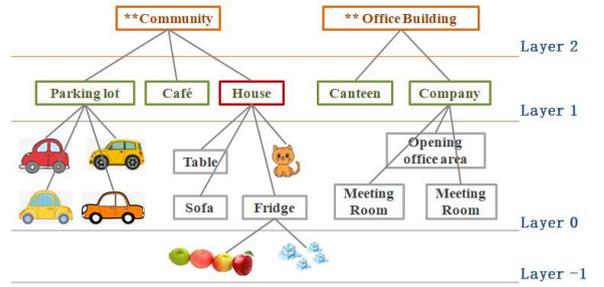

Figure 8. Vertical Tree Structure in SPM

In the tree structure, the spatial scope of the children vertices that own the same parent should be independent of each other; If not, they should be moved up or down until all the children vertices are been independent of each other on the spatial scope.

3) *Typical Expression of Spatial-position Information in Natural Language.:* We can position a target object by deducing its spatial relations with the SAPPs in specific SPM. We use two typical ways to represent the spatial-position of the cat in Fig.7.

a) *Type A expression method:* This is the spatial direction relation recognition method that executes on the graph structure of SPM, which needs to find out the adjacent nodes of the target object on the fixed six directions. See Table X, we respectively take the "Cat" and "Sofa" as target objects to find out their adjacent nodes (SAPP) and the relative direction relation between them on graph structure of SPM and output the sentences. When people (the IPE) use this expression method, they (the IPE) and the target object are usually in the same layer of the SPM.

TABLE X. TYPE A EXPRESSION METHOD OF SPATIAL-POSITION INFORMATION

| Target Object | Spatial-position attribute | | Output sentences |
|---|---|---|---|
| | SAPP | Relative direction (SPW) | |
| Cat | Fridge | Upside (on) | **The** cat *is* on **the** fridge. |
| Sofa | Table | Back side (behind) | **The** sofa *is* behind **the** table. |
| Sofa | Fridge | Left side (on the left side of) | **The** sofa *is* on **the** left side *of* **the** fridge. |

b) *Type B expression method:* This is a spatial scope recognition method that executes on the tree structure of SPM, which needs to find out the parent node of the target object, then use the parent node to assist in positioning the target object. In Fig.8, the parent node of "Apples" is "Fridge", the parent node of "Cat" is "House", and the parent node of "House" is "**community". We use the parent node to define the scope of the target project. People (the IPE) prefer to use this expression method when they (the IPE) are not in the same layer with the target object in SPM.

TABLE XI. TYPE B EXPRESSION METHOD OF SPATIAL-POSITION INFORMATION

| Target Object | Spatial-position attribute | | Output sentences |
|---|---|---|---|
| | *SAPP* | *Scope relation (SPW)* | |
| Apples | Fridge | Inside (in) | **The** apples *are* in **the** fridge. |
| Cat | House | Inside (in) | **The** cat *is* in **the** house. |
| House | **community | Inside (in) | **My** house *is* in **community. |

In general, when people (the IPE) need to describe the position of an object, they need to find out the location of both the object and themselves (IPE) in their SPM, and then to decide which method to choose to express the spatial-positioning information in natural language.

The author just takes the spatial-position AS as an example to elaborate on the spatial attribute information processing mechanism in SPM. Similarly, different attribute information is been processed in different corresponding AS, the corresponding information-processing mechanisms are the foundation of NLU.

B. *Understanding Process of a Dialogue*

Since the author has briefed the new classification of the words and introduced the information process mechanism in one of the spatial-position AS. Now let us put the theory into practice through the example below. The example of a Dialogue and the background information is given in Table XII, and the understanding process is listed in Table XIII.

TABLE XII. DIALOGUE EXAMPLE AND THE BACKGROUND INFORMATION

| Background Information: | Dialogue (CTS [a]: 1st Oct, 17:05): |
|---|---|
| *Jack:* the owner of the house and the home service robot.<br>*Nana:* the home service robot.<br>*Default setting:* the ownership of all the things in the house are belongs to Jack, which means the ownership here can be defined by the spatial attribute. The SPM in Fig.7 is taken as the spatial-position AS in Nana's brain. | *Jack:* "Nana, do we have any apple?"<br>*Nana:* "Yes."<br>*Jack:* "Give me an apple."<br>*Nana:* "Sure." |

a. Timestamps are important tool to identify data creation and termination coordinates on timeline (one of time AS).
CTS: creation timestamp, TTS: termination timestamp.

TABLE XIII. INFORMATION UNDERSTANDING PROCESS ON NANA SIDE

| **Sentence 1:** "Nana, do we have any apple?"<br>**IPE-1**: Jack<br>**IPE-2**: Nana | **Sentence 2:** "Give me an apple."<br>**IPE-1**: Jack<br>**IPE-2**: Nana |
|---|---|
| **Step 1:** Word segmentation | **Step 1:** Word segmentation |
| Nana , do we have any apple ? | Give me an apple . |
| **Step 2:** Identify the sentence pattern | **Step 2:** Identify the sentence pattern |
| Do : Auxiliary word.<br>have: Structure words → Inclusion relation of sets.<br>do…have…? :<br>☐ A-DRM  ☐ A-SRM  ☐ A-PRM<br>☐ B-DRM  ☑ B-SRM  ☐ B-PRM<br>☐ C-DRM  ☐ C-SRM  ☐ C-PRM<br>~~Nana~~ , ~~do~~ ~~we~~ ~~have~~ any apple ? | Give : Data chunk → Verb.<br>Give……[] :<br>☐ A-DRM  ☐ A-SRM  ☑ A-PRM<br>☐ B-DRM  ☐ B-SRM  ☐ B-PRM<br>☐ C-DRM  ☐ C-SRM  ☐ C-PRM<br>~~Give~~ me an apple . |
| **Step 3:** Identify the objects in sentence. | **Step 3:** Identify the objects in sentence. |
| *Object 1:* Jack (we) | *Object 1:* Jack (me) |

| Jack | |
|---|---|
| *Spatial-position* | *Time Position* |
| On sofa | CTS: 1st Oct, 16:30<br>TTS: current time |

*Object 2:* Apple (apple)

| Apple | |
|---|---|
| *Spatial-position* | *Time Position* |
| In fridge (Qty: 3) | CTS: 29th Sep, 11:00<br>TTS: current time |

~~Nana~~ , do we have any ~~apple~~ ?

| Jack | |
|---|---|
| *Spatial-position* | *Time Position* |
| On sofa | CTS: 1st Oct, 16:30<br>TTS: current time |

*Object 2:* Apple (apple)

| Apple | |
|---|---|
| *Spatial-position* | *Time Position* |
| In fridge (Qty: 3) | CTS: 29th Sep, 11:00<br>TTS: current time |

Give ~~me~~ an ~~apple~~ .

**Step 4: Identify the inclusion relation verification task between Object 1 (Jack) and Object 2 (Apple).**

do Object 1 have any Object 2 ? :
Task: to verify whether Object 2 (apple) is a subset of Object 1(Jack).

  Any: Data word → Measuring word.
   Subtask: to verify whether the quantity of Object 2 > 0.

**Step 5: Run the verification tasks, and return the result.**

Task: According to the default setting, all the things in SPM in Fig.7 belongs to Jack. Apple is found in SPM in Fig.7 which means Object 2 is a subset of Object 1
   Return: True

  Subtask: Look up the Memory-sheet of Object 2 in step 3, the quantity of Object 2 = 3 >0.
    Return: True

Return: True (Yes.)
~~Nana , do we have any apple ?~~

All information has been processed
  Continue

---

**Step 4: Identify the action task**

Give Object 1 (sb) Object 2 (sth).
Task: to take a action that to change the spatial-position of Object 2 (apple) from the current position (in fridge) to the position of Object 1 (Jack).

  An: Data word → Measuring word.
   Subtask: give out the quantity of Object 2 that need to be moved.

**Step 5: Take the action** (mobile robot field won't be discussed here)

Return: Achieveable (Sure.)

**Step 6: Update the relevant information in Nana's Memory-sheet.**

| Apple | |
|---|---|
| *Spatial-position* | *Time Position* |
| In fridge (Qty:3) | CTS: 29th Sep, 11:00 |
|  | TTS: 1st Oct, 17:06 |
| In fridge (Qty:2) | CTS: 1st Oct, 17:06 |
|  | TTS: current time |

~~Give me an apple .~~

All information has been processed
  Continue

---

So far, Nana still needs to figure out how to deliver the apple into Jack's hands instead of to deliver it into the sofa. There are still many details to deal with, and the understanding is not entirely precise, but Nana (the home service robot) is already able to understand commands given in natural language.

### IV. THE BASIC CONDITIONS AND DEFINITION OF UNDERSTANDING

We roughly discussed the information architecture through introducing the new classification of lexical and sentences types, which mainly involves the storage structure of data, and the information processing operations implemented on existing information storage database (e.g. information reading, information verification, etc.). It is easy to find out that the same information perception systems, the same information processing systems, and the same information storage database are the basic conditions of NLU. The same physiological structure of human beings ensures that different individuals have the same information perception and processing system. And learning from each other can make up for differences in information storage databases in different brains, thus reducing differences in understanding.

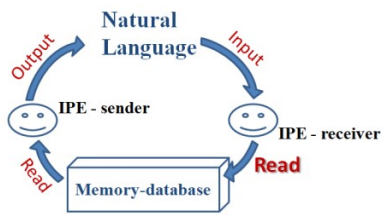

Figure 9. The simplified diagram of the process of reading, transmitting, and understanding of information between information sender and receiver

Now, the author tries to redefine the "understanding" in the following ways:

- At the data level, "Understanding" is mainly about understanding the attributes of EC which include but not limited to the understanding of the variation-rules of their basic attributes, the predictions of the variation-boundaries of their basic attributes in different attribute spaces, the hierarchical relations with other related EC (extended attributes), and the position (advanced attributes) of an EC in the population of a class samples after instantiation.
- At the information-processing level, "Understanding" is about understanding the operational requirements for the target data chunks as expressed in the specific task chunks. Whether it is a data description task, a data verification task, or a data search task, they all involve the following operations: read, write, modify, search, etc.
- IPE always uses their own memory-databases to understand ( or interpret) the input data chunks, the understanding varies when the memory-databases are different. When the input data is entirely new, the IPE needs to learn and build the related data in their memory-database before they can use the memory-database to interpret the input data.

Just like human learning in infancy, the construction of the information memory-database needs to start from the most basic information related to human beings. After having the basic and necessary information memory-database, more abstract information systems such as discipline research and the corresponding knowledge-graph can be built on it. The researches of different disciplines actually are the researches and constructions of specific AS. In this way, all the knowledge in human history can be incorporated into the information memory-database, and been inherited and applied. When this becomes a reality, all human beings will have a shared decision-making system, and humanity will enter a whole new era.